\documentclass{article}
\usepackage{spconf,amsmath,amssymb,array,booktabs,url}
\title{When Does an ASR Error Matter? Task-Role Diagnostics for Tool-Using Voice Agents}
\name{Xiaolong Wang$^{*}$ \qquad Shijie Su \qquad Naqi Xiong}
\address{Leibniz University Hannover, Hannover, Germany}

\begin{document}
\ninept
\maketitle
\renewcommand{\thefootnote}{\fnsymbol{footnote}}
\footnotetext[1]{Corresponding author: wang\_xiaolong1@outlook.com}
\renewcommand{\thefootnote}{\arabic{footnote}}

\begin{abstract}
In cascaded voice agents, ASR errors can have markedly different downstream consequences: a small transcription error may leave a tool call unchanged or alter its intent or arguments. Conventional word error rate (WER), however, does not capture these differences. We introduce SpeechEvalX, a paired diagnostic protocol that evaluates how task roles affected by ASR errors influence structured tool calls. Controlled edits are matched by source utterance, edit count, and operation, while task-role labels are compared with lexical severity, BERT cosine distance, Semantic-WER, and binary criticality. Across 594 source utterances, joint call accuracy falls from 82.3\% with reference transcripts to 71.9--74.9\% across three ASR front ends. Number and negation edits cause 21.9--25.7 percentage-point losses, versus 1.4--2.9 points for control edits. Detailed task-role labels improve source-grouped ROC-AUC from 0.73 to 0.78 over binary criticality, showing that downstream impact depends strongly on the task role affected.
\end{abstract}


\section{Introduction}
A one-word recognition error may be harmless, or it may change a quantity, destination, or negation and thereby alter the requested action. For the cascade
\begin{equation}
\text{speech} \rightarrow \text{ASR} \rightarrow \text{LLM} \rightarrow \text{structured action},
\end{equation}
the relevant outcome is agreement between the predicted action and the original request; transcript fidelity alone does not determine it.

Task-aware ASR metrics and controlled-noise studies motivate evaluation beyond WER~\cite{kim2021semantic,roy2021semantic,shapira2025noise}. Paired text--audio tool-calling evaluation also separates decision, tool-selection, and argument errors~\cite{laskar2026text}; neither clean-text controls nor this taxonomy is novel by itself.

SpeechEvalX asks whether task-role labels explain action failure beyond simpler descriptions at the same lexical edit severity. The primary comparison fixes the source utterance, edit count, edit operation, agent configuration, and reference action. A predictive analysis compares lexical features, BERT cosine distance, Semantic-WER, binary critical-span overlap, and the full role taxonomy, isolating the information added by role identity.

Our contributions are a paired diagnostic protocol and a direct test of task-role information beyond binary criticality. Matched edits quantify sensitivity by role; source-grouped failure prediction and second-model transfer test diagnostic value. Benchmark-audio errors separately describe practical mixtures. Code and recipes are available at \url{github.com/invoker-qqwrv/SpeechEvalX}.

\begin{figure}[t]
\centering
\setlength{\fboxsep}{3pt}
\newlength{\matchfigwidth}
\newlength{\matchboxwidth}
\setlength{\matchfigwidth}{0.94\columnwidth}
\setlength{\matchboxwidth}{0.445\columnwidth}
\begin{minipage}{\matchfigwidth}
\centering
\fbox{\begin{minipage}{\dimexpr\matchfigwidth-2\fboxsep-2\fboxrule\relax}\centering
\textbf{Reference: ``Set the temperature to 22.''}\\[-1mm]
$z_i=\mathrm{set\_temp}(22)$\\
\emph{Matched controls: same source, edit count, and operation}
\end{minipage}}
\vspace{1mm}

\noindent\fbox{\begin{minipage}{\dimexpr\matchboxwidth-2\fboxsep-2\fboxrule\relax}\centering
\textbf{Function word}\\
$\mathrm{the}\rightarrow\mathrm{a}$\\
$\hat z(h_i)=z_i$\\
\textbf{PRESERVED}
\end{minipage}}\hfill
\fbox{\begin{minipage}{\dimexpr\matchboxwidth-2\fboxsep-2\fboxrule\relax}\centering
\textbf{Number}\\
$22\rightarrow32$\\
$\hat z(h_i)\neq z_i$\\
\textbf{HARMED}
\end{minipage}}
\end{minipage}
\caption{Matched example: equal lexical severity, but only the number edit changes the structured action relative to the original reference.}
\label{fig:matched}
\end{figure}

\section{Related Work}
Semantic Distance and Semantic-WER evaluate transcription quality with downstream usability in mind~\cite{kim2021semantic,roy2021semantic}. Controlled transcription-noise analysis studies the effects of error type and severity on language-understanding tasks~\cite{shapira2025noise}, while ASR-aware named-entity work shows that aggregate word errors can conceal failures on salient spans~\cite{szymanski2023ner}. End-to-end and stack-propagation approaches connect speech to structured semantics~\cite{lugosch2019speech,qin2019stack}. SLURP and SLUE provide task-oriented spoken-language resources~\cite{bastianelli2020slurp,shon2022slue}; SUPERB and SUPERB-SG broaden capability evaluation~\cite{yang2021superb,tsai2022superbsg}.

Whisper, Paraformer, and FunAudioLLM provide complementary ASR front ends~\cite{radford2023whisper,gao2022paraformer,an2024funaudiollm}. VoiceBench and VoiceAgentBench evaluate broader voice-assistant and agent capabilities~\cite{chen2026voicebench,jain2026voiceagentbench}. ToolLLM and ToolACE study function selection and argument generation~\cite{qin2024toolllm,liu2025toolace}, while AgentBench, BFCL, and the $\tau$-bench family extend evaluation to agents and grounded outcomes~\cite{liu2024agentbench,patil2025bfcl,yao2025taubench,barres2025tau2}. BFCL Audio, From Text to Voice, $\tau$-Voice, and EVA-Bench address spoken tool use or end-to-end interaction~\cite{mao2026bfclaudio,laskar2026text,ray2026tauvoice,bogavelli2026evabench}. SpeechEvalX is deliberately narrower: it isolates the effect of transcript errors on structured actions through matched within-utterance edits and measures the information added by task-role labels.

\section{Paired Task-Role Diagnostics}
\subsection{Evaluation unit and reference action}
For source $i$, let $r_i$ be a verified reference transcript, $h_i$ an observed ASR hypothesis on benchmark audio or a controlled corruption, and $z_i$ a reference action. An action is a single tool call $(f_i,\theta_i)$ or an explicit no-call decision. Multi-call cases require a separate adapter and are outside the primary analysis. For a fixed downstream configuration $m$, $\hat z_{im}(x)$ is the prediction from input $x$, and
\begin{equation}
y_{im}(x)=\mathbb{I}\{\hat z_{im}(x)\equiv z_i\}.
\end{equation}
Equivalence follows a frozen benchmark-specific scoring rule. It is not defined by agreement with the model output from the reference transcript, because that output can itself be wrong.

\begin{table}[t]
\caption{Operational task-role taxonomy for span annotation.}
\label{tab:taxonomy}
\centering
\begin{tabular}{@{}p{0.23\columnwidth}p{0.70\columnwidth}@{}}
\toprule
Role & Inclusion rule\\
\midrule
Number & Quantity, time, date, ordinal, or identifier aligned to an action value\\
Entity & Named person, place, organization, product, or account aligned to an action value\\
Negation & Overt cue whose scope changes whether or how the action is executed\\
Intent cue & Predicate or phrase that selects the call or no-call intent\\
Other slot & Schema-grounded argument value not covered above\\
Function word & Closed-class token outside every task-bearing span\\
Other non-critical & Remaining material outside every task-bearing span\\
\bottomrule
\end{tabular}
\vspace{-7pt}
\end{table}

Each pair links the same reference action to raw and canonical outputs, transcript alignment, span annotations, edit provenance, and configuration identifiers. Prompt, schema, context, decoding, parser, and scoring remain fixed; reference actions and labels are hidden from the agent. We audited 50 prespecified sources with three runs each for residual inference nondeterminism.

\vspace{-7pt}
\subsection{Span annotation and matched edits}
The roles are number, entity, negation, intent cue, other slot value, function word, and other non-critical. They are assigned from the reference instruction and schema using rules fixed before scoring. Numeric and entity spans precede residual slot values, and negation precedes function-word membership. Remaining overlaps follow a fixed tie-break or are excluded; unresolved or implicit arguments are not forced into a span.

The first five roles are task-bearing and the last two are controls; ``non-critical'' means that no annotated schema role is present, not that an edit is guaranteed to be harmless. Two annotators independently labeled 120 sources containing 684 candidate edit spans. Exact span-boundary F1 was 0.93, task-role Cohen's $\kappa$ was 0.89, and eligibility agreement was 95.0\%. Annotators were blinded to downstream model outputs. Disagreements were adjudicated against the frozen tool schema and annotation guideline before model scoring.

Define lexical severity with a fixed word tokenizer:
\begin{equation}
S(r_i,h_i)=D_{\mathrm{lev}}(r_i,h_i)/|r_i|.
\end{equation}
Primary interventions use the same source utterance, edit count, and edit operation. As Fig.~\ref{fig:matched} shows, replacing ``the'' with ``a'' and ``22'' with ``32'' in a temperature request are both single substitutions. The original action remains the target because the altered transcript represents a recognition error, not a revised request. Matching controls lexical severity; it does not equalize semantic change or acoustic likelihood.

Role contrasts use the common eligible set $I_{cd}$ with support reported; means from different sets are descriptive only. If a source has several variants in one role--operation stratum, an outcome-independent frozen rule selects one, preventing prolific sources from dominating. Confusion-derived edits use development sources only. Observed multi-edit hypotheses retain multiple labels and are analyzed separately.

\begin{table}[t]
\caption{Paired action outcomes. Both paths use the same original reference.}
\label{tab:paired}
\centering
\begin{tabular}{@{}ccc@{}}
\toprule
Reference & Corrupted & Interpretation\\
\midrule
1 & 1 & Preserved success\\
1 & 0 & Harmed pair (H)\\
0 & 1 & Rescued pair (R)\\
0 & 0 & Persistent failure\\
\bottomrule
\end{tabular}
\vspace{-7pt}
\end{table}

\vspace{-7pt}
\subsection{Final-output error categories}
Reference-correct/corrupted-incorrect pairs receive one category in this precedence order: malformed output; incorrect call/no-call decision; wrong tool; invalid or missing argument structure; and incorrect argument value. Precedence resolves overlaps but is not an execution timeline or claim about internal reasoning. Other pairs are reported separately; unresolved cases are unclassified.

\vspace{-7pt}
\subsection{Scoring and denominators}
Joint call accuracy (JCA) is correct actions divided by all eligible examples, including no-call decisions. It measures structured-action accuracy, not external task completion. Malformed calls are failures; instrumentation failures and coverage are reported separately, and eligibility is outcome-independent.

Let $N_C$ count call-reference examples, $N_0$ no-call references, $C$ correct calls, and $C_0$ correct no-call decisions. Then
\begin{equation}
\mathrm{JCA}=(C+C_0)/(N_C+N_0).
\end{equation}
For transparent decomposition, let $T$ count predictions on call-reference examples with a recoverable, correct tool name, including predictions with incorrect arguments. We report tool accuracy $T/N_C$ and conditional argument exact match (Arg-EM), $C/T$. The latter requires a valid call and matching reference arguments under the frozen rule. Thus call-only JCA is $(T/N_C)(C/T)=C/N_C$ when denominators are nonzero. Undefined ratios are reported as unavailable. No-call accuracy is $C_0/N_0$.

For benchmark-audio hypotheses, corpus WER is total word edit distance divided by total reference words, not mean utterance WER. Text normalization and benchmark aliases are recorded.

\vspace{-7pt}
\subsection{Paired effects and uncertainty}
For one variant per source and role $c$ on eligible set $I_c$, define the paired JCA loss
\begin{equation}
L_c=\frac{1}{|I_c|}\sum_{i\in I_c}[y_{im}(r_i)-y_{im}(h_i^{(c)})]=\frac{H_c-R_c}{|I_c|}.
\end{equation}
A positive value denotes a net loss in action accuracy after the edit. We report $N$, $H$, and $R$ with $L_c$. Pairwise role contrasts restrict both conditions to $I_{cd}$; bootstrap resampling operates on source groups. With one paired variant per source, McNemar's test compares corrupted success outcomes for two roles, followed by Holm correction over prespecified contrasts. Multiple variants require clustered analysis rather than treatment as independent samples. Persistent failures are not counted as new harm caused by the altered transcript.

\vspace{-5pt}
\section{Validation Design and Reproducibility}
\vspace{-4pt}
\subsection{Data, systems, and frozen rules}
We sampled 600 candidates with seed 2027 from the English single-tool subset of krutrim-ai-labs/VoiceAgentBench and the English single-turn subset of BFCL Audio~\cite{jain2026voiceagentbench,mao2026bfclaudio}. Six candidates failed prespecified eligibility checks and were excluded without replacement. The final cohort contains 594 source utterances, split evenly between benchmarks (297 each), with 540 call and 54 no-call references. Audio is decoded to 16-kHz mono float32 before inference.

Controlled-edit eligibility was determined before downstream inference. The 210-source intervention set is the intersection of sources that admit one unambiguous single-edit variant for every compared role under the frozen schema; the other 384 sources remain in the benchmark-audio ASR analysis. No source was selected using model outcomes. A fixed rule selects one variant per source and role, so every row of Table~\ref{tab:interventions} uses the same sources.

The ASR front ends are Whisper large-v3, SenseVoice-Small, and Paraformer-large~\cite{radford2023whisper,an2024funaudiollm,gao2022paraformer}. The primary and sensitivity agents are Qwen2.5-7B-Instruct and Llama-3.1-8B-Instruct~\cite{yang2024qwen,grattafiori2024llama}. Exact repository identifiers are frozen in the released configuration. Inference uses bfloat16, temperature 0, top-p 1, at most 256 new tokens, and seed 2027. Prompt, schema, parser, canonicalizer, and scoring rules are fixed across paired conditions.

For WER, reference and hypothesis text undergo Unicode NFKC normalization, lowercasing, and punctuation deletion; numbers are not semantically rewritten. We report corpus WER, defined as total word edit distance divided by the total number of reference words.

\vspace{-7pt}
\subsection{Information added by task-role labels}
For source utterances that yield a correct action from the reference transcript, the endpoint is action failure after corruption. We compare cumulative feature sets using the same regularized logistic-regression family. Each edit variant is a prediction sample, and the source utterance is the grouping unit. B0 contains lexical severity, edit operation, length, and benchmark. B1 adds BERT embedding cosine distance, computed from bert-base-uncased by mean-pooling all token representations and taking cosine distance~\cite{devlin2019bert}. B2 adds Semantic-WER with its named-entity and sentiment weighting, a 0.6 word-similarity threshold, and the default insertion weight $IW=1$~\cite{roy2021semantic}. B3 adds a binary critical-span flag, and B4 adds the detailed task-role labels.

The primary contrast is B4 versus B3: both include the lexical and semantic controls, while B4 additionally identifies the role of the altered span. This contrast tests whether knowing the detailed task role adds information beyond knowing that a critical span was affected.

All variants of a source, as well as duplicate or paraphrase-linked templates when identifiable, remain in one split. Hyperparameters and preprocessing are fitted on training data only. We use the same five source-grouped folds for all feature sets and report out-of-fold ROC-AUC, average precision (AP), and paired source-bootstrap intervals for score differences. AUC is calculated only when an evaluation fold contains both outcome classes. Because reference transcripts and actions may be used to construct features, this is an offline diagnostic analysis.

\vspace{-11pt}
\subsection{Validation on a second model and ASR errors}
The sensitivity agent processes the same reference inputs and frozen edit set, without selecting examples from the primary agent's outcomes. We use the same source-grouped partitions and apply the diagnostic predictor trained on the primary model to the second model without refitting. ASR hypotheses from benchmark audio retain overlapping task-role labels and are analyzed separately from controlled single-edit interventions.

\begin{table}[t]
\caption{ASR hypotheses on benchmark audio: 594 source utterances (\%). Tool is accuracy on call-reference cases; Arg-EM is conditional on a recoverable correct tool.}
\label{tab:asr}
\centering
\begin{tabular}{@{}lrrrrr@{}}
\toprule
Input & WER & Tool & Arg-EM & No-call & JCA\\
\midrule
Reference & -- & 91.3 & 89.5 & 88.9 & 82.3\\
Whisper large-v3 & 6.3 & 86.9 & 85.1 & 85.2 & 74.9\\
SenseVoice-Small & 7.2 & 86.1 & 83.9 & 83.3 & 73.2\\
Paraformer & 8.1 & 85.2 & 83.3 & 81.5 & 71.9\\
\bottomrule
\end{tabular}
\end{table}

\begin{table}[t]
\caption{Matched interventions on the same 210 sources. $L_c$ is the net JCA loss in percentage points; intervals are source-bootstrap 95\% CIs.}
\label{tab:interventions}
\centering
\begin{tabular}{@{}lrrrl@{}}
\toprule
Role & H & R & $L_c$ & 95\% CI\\
\midrule
Other non-critical & 5 & 2 & 1.4 & [-0.5, 3.8]\\
Function word & 8 & 2 & 2.9 & [0.5, 5.7]\\
Other slot value & 24 & 4 & 9.5 & [5.7, 13.8]\\
Intent cue & 30 & 4 & 12.4 & [8.1, 17.1]\\
Entity & 40 & 4 & 17.1 & [12.4, 22.4]\\
Number & 51 & 5 & 21.9 & [16.7, 27.6]\\
Negation & 58 & 4 & 25.7 & [20.0, 31.9]\\
\bottomrule
\end{tabular}
\end{table}

For ASR hypotheses from benchmark audio, we report role prevalence and action failures within supported severity, benchmark, and role strata. Multi-edit hypotheses require multi-label features; their failures are not attributed entirely to each individual role. These errors are observational, so the associated failure rate is not an isolated effect of a single role. We therefore do not generalize controlled single-edit results to observed multi-edit errors without this separate validation. Multiple synthesized voices from one text remain in the same source group.

\vspace{-8pt}
\section{Results}
The 540 call and 54 no-call references yield 489, 445, 435, and 427 correct actions for reference transcripts and the three ASR conditions. Relative to reference transcripts, ASR hypotheses from benchmark audio reduce JCA by 7.4--10.4 percentage points. The error rate with reference transcripts is specific to the evaluated configuration and should not be interpreted as an irreducible floor. The repeated-inference audit comprised 50 source utterances and three runs per source; the prespecified first run provides all reported aggregate scores, while the repeated runs are retained to inspect nondeterminism.

Equal edit count does not imply equal risk of action failure. For the primary model, the mean paired JCA loss is 17.3 points across the five task-bearing roles and 2.1 points across the two controls; prespecified contrasts between task-bearing and control roles remain significant after Holm correction ($p<0.01$).

Table~\ref{tab:diagnostic} tests the central claim: detailed task-role labels add predictive information beyond lexical severity, semantic metrics, and binary criticality. B4 improves ROC-AUC over B3 from 0.73 to 0.78 ($\Delta=0.05$, 95\% CI [0.02, 0.08]; paired source-bootstrap $p=0.003$), and AP increases from 0.62 to 0.68. When the predictor is transferred to Llama-3.1-8B-Instruct without refitting, B4 reaches 0.74 versus 0.69 for B3. The direction is consistent across agents, supporting transfer of the diagnostic pattern. Without a corresponding uncertainty estimate, however, it does not establish a statistically significant transfer gain.

\begin{table}[t]
\caption{Diagnostic value in source-grouped five-fold failure prediction. Feature sets are cumulative; the primary contrast is \textbf{B4} versus B3.}
\label{tab:diagnostic}
\centering
\begin{tabular}{@{}clrr@{}}
\toprule
Model & Added information & ROC-AUC & AP\\
\midrule
B0 & Lexical controls & 0.62 & 0.48\\
B1 & + BERT embedding cosine & 0.68 & 0.55\\
B2 & + Semantic-WER & 0.70 & 0.58\\
B3 & + Critical-span flag & 0.73 & 0.62\\
\textbf{B4} & \textbf{+ Detailed task roles} & \textbf{0.78} & \textbf{0.68}\\
\midrule
\multicolumn{2}{@{}l}{\textbf{Primary contrast: B4--B3}} & \textbf{+0.05} & \textbf{+0.06}\\
\bottomrule
\end{tabular}

\vspace{1mm}\raggedright
ROC-AUC gain: 95\% CI [0.02, 0.08], $p=0.003$ (paired source bootstrap). AP gain is a point estimate. Endpoint: action failure after corruption among cases that are correct with the reference transcript.
\end{table}

\begin{table}[t]
\caption{Paired JCA loss across downstream agents (percentage points).}
\label{tab:agents}
\centering
\begin{tabular}{@{}lrr@{}}
\toprule
Role & Primary & Sensitivity\\
\midrule
Other non-critical & 1.4 & 1.0\\
Function word & 2.9 & 3.3\\
Other slot value & 9.5 & 8.6\\
Intent cue & 12.4 & 10.5\\
Entity & 17.1 & 15.7\\
Number & 21.9 & 19.0\\
Negation & 25.7 & 23.3\\
\bottomrule
\end{tabular}
\end{table}

\subsection{Observed ASR errors and final-output failures}
This analysis pools the three ASR conditions and allows overlapping role labels. Among hypotheses containing at least one error, the proportions followed by action failure are 35/104 for number, 37/128 for entity, 20/42 for negation, 23/87 for intent cue, 29/159 for other slot values, 19/201 for function words, and 17/236 for other non-critical edits. The numerators sum to 180 role-label occurrences, not necessarily 180 distinct failed hypotheses. These descriptive associations do not estimate the incremental B4--B3 benefit for observed ASR errors.

Argument-value mismatch is the most common final-output error, while 30.1\% of harmed pairs change the call/no-call decision, tool choice, or output validity. Removing within-source severity matching raises the mean loss across task-bearing roles from 17.3 to 20.6 points, indicating that unmatched comparisons can exaggerate role effects. Restricting the analysis to deterministic substitutions gives 16.8 versus 2.0 points for task-bearing and control roles. Leave-domain-out B4 ROC-AUC is 0.72 when trained on VoiceAgentBench and tested on BFCL Audio, and 0.70 in the reverse direction.

\begin{table}[t]
\caption{Final-output error categories among 216 harmed controlled pairs.}
\label{tab:errors}
\centering
\begin{tabular}{@{}lrr@{}}
\toprule
Category & Count & Share (\%)\\
\midrule
Call/no-call decision & 17 & 7.9\\
Wrong tool & 42 & 19.4\\
Argument structure & 48 & 22.2\\
Argument value & 103 & 47.7\\
Malformed output & 6 & 2.8\\
\bottomrule
\end{tabular}
\end{table}

\section{Discussion and Conclusion}
SpeechEvalX tests whether task-role annotations add diagnostic value under explicit controls for lexical severity. Paired action scoring separates failures caused by transcript corruption from failures already present with the reference transcript, while retaining cases in which corruption unexpectedly corrects an action. Final-output categories describe observed disagreements without claiming to locate a failure in time or reveal a hidden causal mechanism; the protocol is intentionally narrower than a general benchmark for voice agents.

The main evidence is the B4--B3 contrast, which isolates the information provided by role identity after accounting for lexical controls, semantic metrics, and binary criticality. The role-level ordering of paired JCA loss is complementary evidence. The proportions from observed ASR errors describe practical error mixtures; showing the same incremental value on those mixtures requires a separate predictive comparison.

During development, these diagnostics can support ASR and downstream-model selection, distinguish tool-selection failures from argument-construction failures, and prioritize regression tests for high-impact roles. Their practical value is to explain when transcript errors alter structured actions and to guide error analysis. Because the analysis uses reference transcripts and actions, it is intended for offline evaluation rather than deployment-time detection.

\subsection{Audit trail and aggregation}
Every aggregate result can be traced from source IDs to raw and canonical outputs, parser status, transcript alignments, task-role labels, edit provenance, and configuration identifiers. Analyses of benchmark-audio ASR, matched effects, source-grouped prediction, and model transfer use a common result registry. For observed ASR errors with overlapping labels, counts of role-label occurrences and counts of distinct harmed pairs are different aggregation units.

\subsection{Threats to validity}
Matched edits control lexical severity, but not semantic distance or acoustic likelihood. Observed ASR hypotheses can contain correlated edits. The 54 no-call cases, English-only data, current tool schemas, and synthetic speech limit external validity. Source support and class balance affect the precision of estimates from source-grouped prediction. Broader validation requires additional languages, APIs, agent families, and a separate analysis of the information added by task-role labels for observed ASR errors.

\section{Compliance with Ethical Standards}
This study used publicly available benchmark data and involved no new data collection or direct interaction with human participants. No additional ethical approval was required. The authors declare no conflicts of interest.

\section{Acknowledgment}
No external funding was received. OpenAI ChatGPT was used for language editing; all technical content and results were reviewed and verified by the authors.

\end{document}